\documentclass[runningheads]{llncs}

\usepackage[T1]{fontenc}

\usepackage{graphicx}
\usepackage{subfigure}

\usepackage{amsmath}
\usepackage{amssymb}
\usepackage{mathtools}

\usepackage{bbm}
\usepackage{booktabs, multirow, makecell} 

\usepackage{hyperref}
\usepackage{color}

\makeatletter
\newcommand\appendix@section[1]{%
  \refstepcounter{section}%
  \orig@section*{Appendix \@Alph\c@section: #1}%
  \addcontentsline{toc}{section}{Appendix \@Alph\c@section: #1}%
}
\let\orig@section\section
\g@addto@macro\appendix{\let\section\appendix@section}
\makeatother

\begin{document}

\title{Defending Observation Attacks in Deep Reinforcement Learning via Detection and Denoising}
\titlerunning{Defending Obs. Attacks via Detection and Denoising}

\author{Zikang Xiong\inst{1} \and
Joe Eappen\inst{1} \and
He Zhu\inst{2} \and 
Suresh Jagannathan\inst{1}
}

\authorrunning{Z. Xiong et al.}
\institute{Purdue University, West Lafayette IN 47906, USA \and
Rutgers University, New Brunswick NJ, 08854, USA\\
\email{{xiong84,jeappen}@purdue.edu, \\ hz375@cs.rutgers.edu, suresh@cs.purdue.edu}}

\maketitle              % typeset the header of the contribution

\begin{abstract}
    Neural network policies trained using Deep Reinforcement Learning (DRL) are well-known to be susceptible to adversarial attacks. In this paper, we consider attacks manifesting as perturbations in the observation space managed by the external environment. These attacks have been shown to downgrade policy performance significantly. We focus our attention on well-trained deterministic and stochastic neural network policies in the context of continuous control benchmarks subject to four well-studied observation space adversarial attacks. To defend against these attacks, we propose a novel defense strategy using a detect-and-denoise schema. Unlike previous adversarial training approaches that sample data in adversarial scenarios, our solution does not require sampling data in an environment under attack, thereby greatly reducing risk during training. Detailed experimental results show that our technique is comparable with state-of-the-art adversarial training approaches.
\end{abstract}

\section{Introduction}
\label{sec:intro}

Deep Reinforcement Learning (DRL) has achieved promising results in many challenging continuous control tasks. However, DRL controllers have proven vulnerable to adversarial attacks that trigger performance deterioration or even unsafe behaviors. For example, the operation of an unmanned aerial navigation system may be degraded or even maliciously affected if the training of its control policy does not carefully account for observation noises introduced by sensor errors, weather, topography, obstacles, etc. Consequently, building robust DRL policies remains an important ongoing challenge in architecting learning-enabled applications.

There have been several different formulations of DRL robustness that have been considered previously. \cite{mandlekar2017adversarially,rajeswaran2016epopt} consider DRL robustness against perturbations of physical environment parameters. More generally, \cite{iyengar2005robust} has formalized DRL robustness against uncertain state transitions, and \cite{tessler2019action} has studied DRL robustness against action attacks. Similar to \cite{zhang2020robust}, our work considers DRL robustness against \emph{observation attacks}. Prior work has demonstrated a range of strong attacks in the observation space of a DRL policy~\cite{lin2017tactics,pattanaik2017robust,kos2017delving,zhang2020robust,zhang2021robust,sun2020stealthy}, all of which can significantly reduce a learning-enabled system's performance or cause it to make unsafe decisions. Because observations can be easily perturbed, robustness to these kinds of adversarial attacks is an important consideration that must be taken into account as part of a DRL learning framework. There have been a number of efforts that seek to improve DRL robustness in response to these concerns. These include enhancing DRL robustness by adding a regularizer to optimize goals~\cite{achiam2017constrained,zhang2020robust} and defending against adversarial attacks via switching policies~\cite{havens2018online,xiong2020robustness}. There have also been numerous proposals to improve robustness using adversarial training methods. These often require sampling observations under \emph{online} attacks (e.g., during simulation) \cite{kos2017delving,pattanaik2017robust,zhang2021robust}.  However, while these approaches provide more robust policies, it has been shown that such approaches can negatively impact policy performance in non-adversarial scenarios. Moreover, a large number of unsafe behaviors may be exhibited during online attacks, potentially damaging the system controlled by the learning agent if adversarial training takes place in a physical rather than simulated environment.

\begin{figure}[t]
    \includegraphics[width=0.92\linewidth]{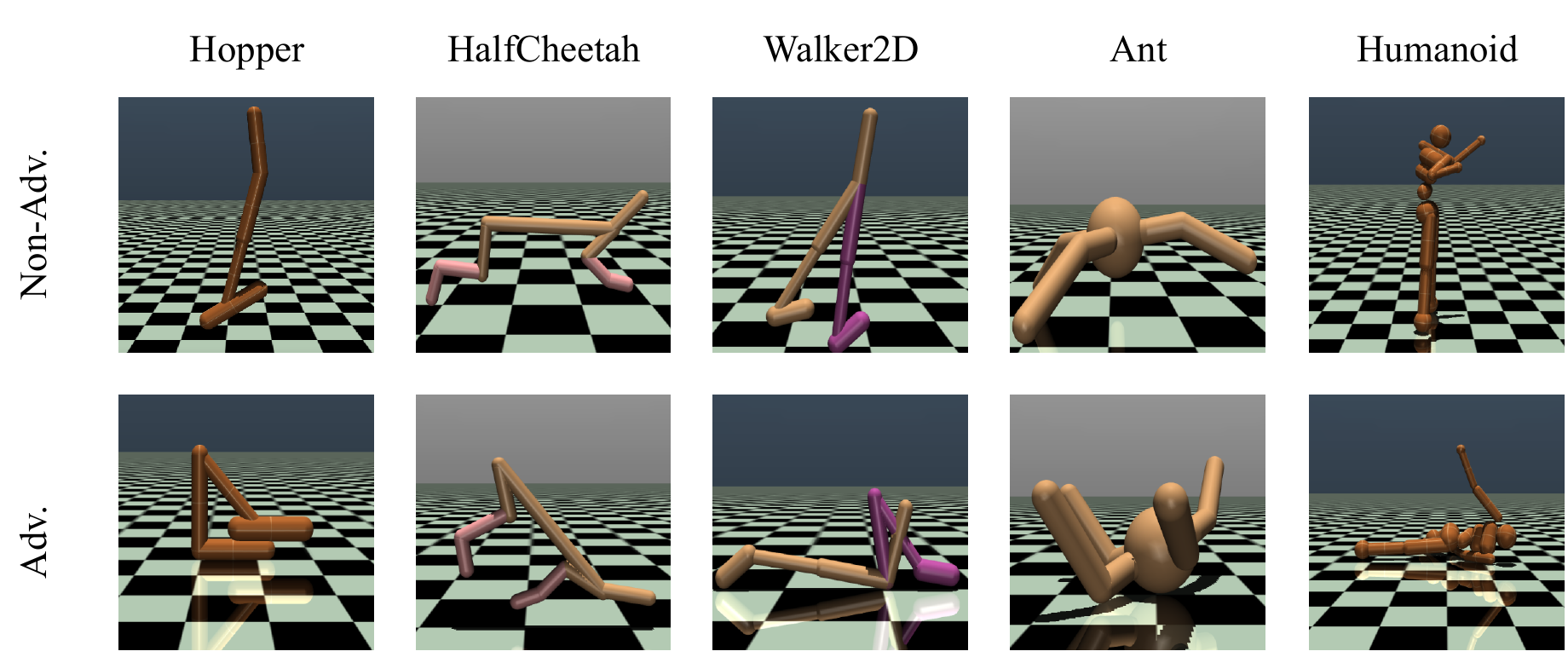}
    \caption{Robots we evaluated in non-adversarial and adversarial scenarios. Robots fall down and gain less rewards when they are under attack.}
    \label{fig:agents-demonsrtation}
\end{figure}

\begin{figure}[ht]
    \centering
    \includegraphics[width=0.9\linewidth]{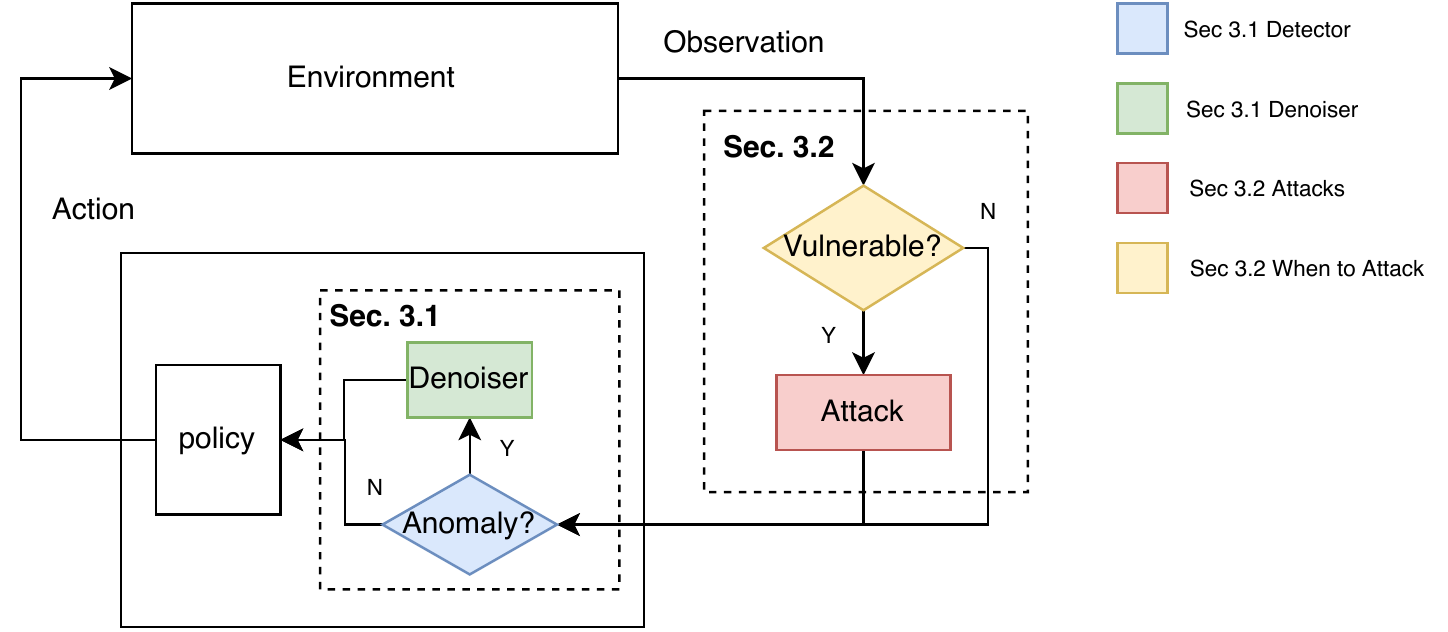}
    \caption{Framework}
    \label{fig:framework}
\end{figure}

To address the aforementioned challenges, we propose a new algorithm that strengthens the robustness of a DRL policy \emph{without} sampling data under adversarial scenarios, avoiding the drawbacks that ensue from encountering safety violations during an online training process. Our method is depicted in Fig.~\ref{fig:framework}. Given a DRL policy $\pi$, our defense algorithm \emph{retains} $\pi$'s parameters and trains a \emph{detector} and \emph{denoiser} with offline data augmentation. The detector and denoiser address problems on when and how to defend against an attack, resp. When defending $\pi$ in a possibly adversarial environment, the detector identifies anomalous observations generated by the adversary, and the denoiser processes these observations to reverse the effect of the attacks. With assistance from the detector and the denoiser, the algorithm overcomes adversarial attacks in the policy's observation space while retaining performance in terms of the achieved total reward.

Both the detector and denoiser are modeled with Gated Recurrent Unit Variational Auto-Encoders (GRU-VAE). This design choice is inspired by recent work \cite{malhotra2016lstm,zhang2019unsupervised,vincent2008extracting,park2018multimodal,su2019robust} that has demonstrated the power of such anomaly detectors and denoisers. After anomalies enforced by attacks are detected, we need to reverse the effect of the attacks with a denoiser. However, training such a denoiser requires the observations under attack as input, but sampling such adversarial observations online is unappealing. To avoid unsafe sampling, our algorithm instead conducts adversarial attacks using offline data augmentation on a dataset of observations collected by the policy in a non-adversarial environment.

Our approach provides several important benefits compared with previous online adversarial training approaches. First, because we do not retrain victim policies, our approach naturally \emph{retains} a policy's performance in non-adversarial scenarios. Second, unlike adversarial training methods that need to sample data under online adversarial attacks, we only require sampled observations with a pretrained policy in a normal environment not subject to attacks. Third, the stochastic components in our detect-and-denoise pipeline (i.e., the prior distribution in the variational autoencoders) provide a natural barrier to defeat adversarial attacks~\cite{panda2018implicit,li2018certified}. We have evaluated our approach on a range of challenging MuJoCo~\cite{mujoco} continuous control tasks for both deterministic TD3 policies~\cite{fujimoto2018addressing} and stochastic PPO policies~\cite{schulman2017proximal}. Our experimental results show that compared with the state-of-the-art online adversarial training approaches~\cite{zhang2021robust}, our algorithm does not compromise policy performance in perturbation-free environments and achieves comparable policy performance in environments subject to adversarial attacks.

To summarize, our contributions are as follows:

\begin{itemize}
    \item We integrate autoencoder-style anomaly detection and denoising into a defense mechanism for DRL policy robustness and show that the defense mechanism is effective under environments with strong known attacks as well as their variants and does not compromise policy performance in normal environments.
    \item We propose an adversarial training approach that uses offline data augmentation to avoid risky online adversarial observation sampling.
    \item We extensively evaluate our defense mechanism for both deterministic and stochastic policies using four well-studied categories of strong observation space adversarial attacks to demonstrate the effectiveness of our approach.
\end{itemize}

\section{Background}

\subsection{Markov Decision Process}

A Markov Decision Process (MDP) is widely used for modeling reinforcement learning problems. It is described as a tuple $(S, A, T, R, \gamma, O, \phi)$. $S$ and $A$ represent the state and action space, resp. $T(s, a): S \times A \rightarrow \mathbbm{P}(S)$ is the transition probability distribution. Given current state $s$ and the action $a$, the Markov probability transition function $T(s, a)$ returns the probability of a new state $s'$. $R(s, a, s'): S \times A \times R \rightarrow \mathbb{R}$ is the reward function that measures the performance of a given transition $(s, a, s')$. Let the cumulative discounted reward be $\mathcal{R}$, and the reward at time $t$ be $R(s_t, a_t, s_{t+1})$.  Then, $\mathcal{R} = \sum_{t=0}^{T} \gamma^t R(s_t, a_t, s_{t+1})$, where $\gamma \in [0, 1)$ is the discounted factor and $T$ is the maximum time horizon. The last element in the MDP tuple is an observation function $\phi: S \rightarrow O$ which transforms states in the state space $S$ to the observation space $O$.  The task of solving an MDP is tantamount to finding an optimal policy $\pi: O \rightarrow A$ that maximizes the discounted cumulative reward $\mathcal{R}$.

\subsection{Observation Attack}

Given a pretrained policy $\pi$, the observation attack $\mathcal{A}_\mathcal{B}$ injects noise to the observation to downgrade the cumulative reward $\mathcal{R}$.  $\mathcal{B}$ quantifies this noise term. Typically, $\mathcal{B}$ is an $\ell_n$-norm region around the ground-truth observation. Given an observation $o_t$, $\mathcal{B}(o_t) = \{\hat{o}_t \mid \ ||\hat{o}_t - o_t||_n < \varepsilon \}$, where $\varepsilon$ is the radius of the $\ell_n$ norm region.  Additionally, attacks can choose when to inject noise. Since it is crucial to downgrade performance using as few attacks as possible, it is typical to define a vulnerability indicator $\mathbbm{1}_{vul}: O \rightarrow \{\mathtt{True}, \mathtt{False}\}$. Given an observation $o_t$, if $\mathbbm{1}_{vul}(o_t)$ is $\mathtt{False}$, the policy $\pi$ receives the perturbation-free observation $o_t$ as input; otherwise, the input will be an adversarial observation $\hat{o}_t = \mathcal{A}_\mathcal{B}(o_t)$.

\subsection{Defense via Detection and Denoising}

The MDP tuple becomes $(S, A, T, R, \gamma, O, \phi, \mathcal{A}_\mathcal{B}, \mathbbm{1}_{vul})$ after incorporating an adversary.  One way to defend against adversarial attacks is to retrain a policy for the new MDP. However, such an approach ignores the fact that we already have a trained policy that performs well in non-adversarial scenarios. Additionally, the solution of such an MDP may not yield an optimal policy \cite{zhang2020robust}. In contrast, our approach considers removing the effects introduced by $\mathcal{A}_\mathcal{B}, \mathbbm{1}_{vul}$ by casting the adversarial MDP problem back into a standard MDP. To do so, we exploit the trained policy and avoid the possibility of failing to find an optimal policy, even in non-adversarial scenarios. Notably, our approach eliminates the effect introduced by the vulnerability indicator $\mathbbm{1}_{vul}$ and observation attack $\mathcal{A}_\mathcal{B}$ by using an anomaly detector and a denoiser, resp. Given a sequence of observations $h_t = \{o_0, ..., o_t\}$, the detector is tasked with predicting whether an attack happens in the latest observation $o_t$. Conversely, the denoiser predicts the ground-truth observation of $o_t$ with $h_t$. If the detector finds an anomaly, the denoiser's prediction is used to replace the current observation $o_t$ with the ground-truth observation. Our defense only intervenes when the detector reports an anomaly, which preserves the performance of pretrained policies when no adversary appears. Training a VAE denoiser typically requires both the groundtruth inputs (i.e., the actual observations) and the perturbed inputs (i.e., the adversarial observations). However, sampling the adversarial observations under online adversarial attacks can be risky. Thus, we prefer sampling adversarial observations offline.

\subsection{Online and Offline Sampling}

The difference between online and offline sampling manifests in whether we need to sample data via executing an action in an environment. Adversarial attacks can downgrade performance by triggering unsafe behaviors (e.g., flipping an ant robot, letting a humanoid robot fall), and hence online sampling adversarial observations can be risky. In contrast, offline sampling does not collect data via executing actions in an environment and thus does not suffer from potential safety violations when performing the sampling online. Here, adversarial observations are sampled offline by running adversarial attacks on a normal observation dataset (i.e., observations generated in non-adversarial scenarios).

\section{Approach}
The overall framework of our approach is shown in Fig.~\ref{fig:framework}. Our defense technique is presented in Section~\ref{sec:defense}. It consists of two components: a detector and a denoiser. First, the anomaly detector checks whether the current environment observation is an anomaly due to an adversarial attack. When an anomaly is detected, the denoiser reverses the attack by denoising the perturbed observation. We evaluate our defense strategy over four attacks described in Section~\ref{sec:attack}. Similar to \cite{lin2017tactics,kos2017delving}, our framework allows an adversary, when given an observation, to decide whether the observation is vulnerable to an attack. 
% Compared with attacking a policy at every step or randomly, restricting attacks to only vulnerable observations enables attackers to deteriorate reward performance in fewer steps. \ZX{Deleting because we did not evaluate that the attack time tiggered. Feel like that we just need to mention this is a common practice in previous work}

\subsection{Defense}
\label{sec:defense}

Adversarial training has broad applications to improve the robustness of machine learning models by augmenting the training dataset with samples generated by adversarial attacks. In the context of deep reinforcement learning, previous approaches~\cite{mandlekar2017adversarially,kos2017delving,pattanaik2017robust,zhang2021robust} conduct policy searches in environments subject to such attacks, leading to robust policies under observations generated from adversarial distributions. As mentioned earlier, our method is differentiated from these approaches by using a detect-and-denoise schema learnt from offline data augmentation while keeping pretrained policies.

Prior work has shown that the LSTM-Autoencoder structure outperforms other methods in various anomaly settings \cite{malhotra2016lstm,zhang2019unsupervised,vincent2008extracting} including anomaly detection in real-world robotic tasks \cite{park2018multimodal}. Inspired by the success of this design choice, we choose to implement both the detector and denoiser as Gated Recurrent Unit Variational Auto-Encoders (GRU-VAE). 

\begin{figure}[ht]
    \centering
    \includegraphics[width=0.8\linewidth]{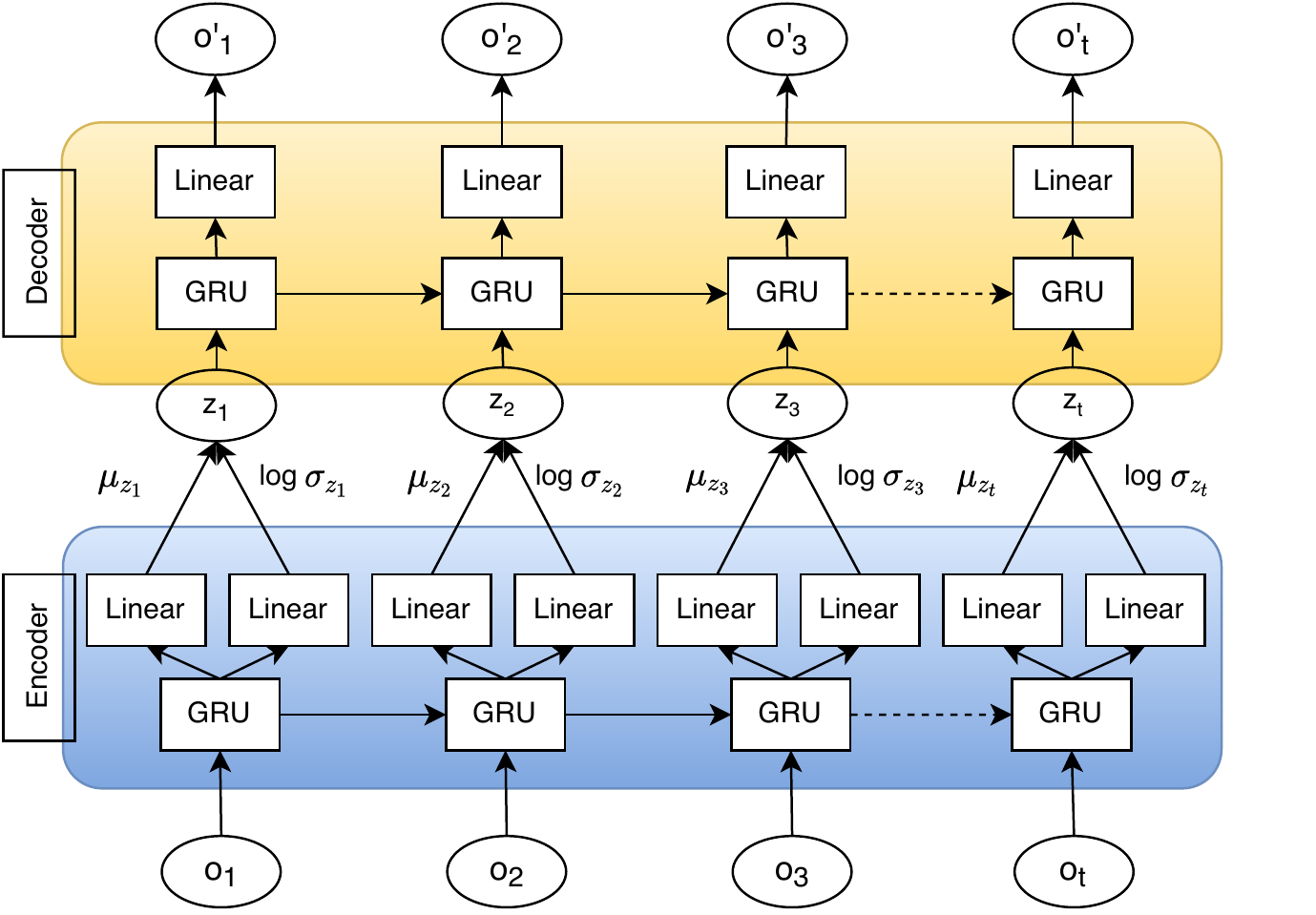}
    \caption{GRU-VAE. The input of the encoder is a sequence of observations. These observations pass a GRU layer and two different linear layers to generate the mean $\mu_t$ and log variance $log\ \sigma_{t}$ of a Gaussian distribution. The latent variable $z_t$ is sampled from this Gaussian distribution, and is passed to the decoder. The decoder decodes $z_t$ with a GRU and a linear layer. The decoder is a deterministic model. For the detector, the output of the decoder is trained to be the same as the input observation sequence. For the denoiser, the output is trained to remove perturbations injected by adversaries. }
    \label{fig:gru-vae}
\end{figure}

\subsubsection{Detector}
The structure of our detector is depicted in Figure~\ref{fig:gru-vae}. The detector learns what normal observation sequences should be. We train it with an observation dataset $\mathcal{D}_{normal}$ sampled online with a pretrained policy in non-adversarial environments. The objective function is the standard variational autoencoder lower bound~\cite{doersch2016tutorial},
\begin{align*}
    L_{\mathtt{det}}= \mathbb{E}_{q_{\theta_q}(z_t \mid o_t, h^{o}_t)}\left[\log p_{\theta_p}(o_t \mid z_t, h^{z}_t)\right] - D_{KL}\left(q_{\theta_q}(z_t \mid o_t, h^{o}_t) \| pr(z_t)\right)
\end{align*}
where $\theta_q$ is the parameters of encoder $q_{\theta_q}$ and $\theta_p$ is the parameters of decoder $p_{\theta_p}$; $o_t \in \mathcal{D}_{normal}$ is the observation at time $t$; $h^{o}_t$ and $h^{z}_t$ are the hidden states for the encoder and decoder, resp.; and $z_t$ is sampled from the distribution parameterized by $q_{\theta_q}$.  Decoding the latent variable $z_t$ reconstructs the input observation $o_t$. $\mathbb{E}_{q_{\theta_q}(z_t \mid o_t, h^{o}_t)}\left[\log p_{\theta_p}(o_t \mid z_t, h^{z}_t)\right]$ is known as the reconstruction objective, the maximization of which increases the likelihood of reconstructing the observations sampled by the pre-trained policy. $D_{KL}\left(q_{\theta_q}(z_t \mid o_t, h^{o}_t) \| pr(z_t)\right)$ is the KL-divergence between the distribution $q_{\theta_q}(z_t \mid o_t, h^{o}_t)$ generated by the encoder and the prior distribution $pr(z_t)$, which serves as the KL regularizer that makes these two distributions similar. Following~\cite{park2018multimodal}, we set $pr(z_t)$ as a Gaussian distribution whose covariance is the identity matrix $I$, but leave the mean of $pr(z_t)$ to be $\mu_{z_t}$ instead of 0.  
The learnable $\mu_{z_t}$ allows the mean of the prior distribution to be conditioned on input observations. This modified GRU-VAE is different from a general GRU-VAE model which assumes the prior distribution is a fixed normal distribution. It depends on the decoder to provide prior distributions, which is crucial for a detector as shown in ~\cite{park2018multimodal}.

The detector reports an anomaly observation when a decoded observation is significantly different from the encoded observation, measured by the $\ell_\infty$-norm between the input observation $o_t$ and the output observation $o'_t$. The detector reports the anomaly if the $\ell_\infty$-norm between $o_t$ and $o'_t$ is greater than a threshold $C_{anomaly}$ using by the anomaly detection indicator function:
\begin{align*}
    \mathbbm{1}_{anomaly}(o_t, o'_t) := \ell_\infty(o_t, o'_t) > C_{anomaly}
\end{align*}

\subsubsection{Denoiser}
The denoiser learns to map anomaly observations found by the detector to the ground-truth normal observations. The objective function of the denoiser is:
\begin{align*}
    L_{\mathtt{den}}= \mathbb{E}_{{q'}_{\theta_{q'}}(z_t \mid \bar{o}_t, h^{\bar{o}_t})}\left[\log {p'}_{\theta_{p'}}(o_t \mid z_t, h^{z}_t)\right] - D_{KL}\left({q'}_{\theta_{q'}}(z_t \mid \bar{o}_t, h^{\bar{o}}_t) \| pr(z_t)\right)
\end{align*}
Compared to the detector's objective function $L_{\mathtt{det}}$, the input to the encoder ${q'}_{\theta_{q'}}$ is replaced by an observation $\bar{o}_t \in \mathcal{D}_{adv} \cup \mathcal{D}_{normal}$ and hidden state $h^{\bar{o}}_t$. The encoder of the denoiser maps $\bar{o}_t$ and hidden state $h^{\bar{o}}_t$ to a latent variable $z_t$ that is used by the decoder ${p'}_{\theta_{p'}}$ to generate the ground-truth observation $o_t$. We also leave the mean of $pr(z_t)$ to be $\mu_{z_t}$ as in the detector. 
Training the denoiser requires the observation $\bar{o}_t$, which we have to sample by conducting adversarial attacks.

Sampling adversarial observations online is generally viewed to be a costly requirement because it must handle potentially unsafe behaviors that might manifest; these behaviors could damage physical agents (e.g., robots) during training (e.g., by causing a robot to fall down). In contrast, we generate these adversarial observations offline. First, since a well-trained policy already exists, we can sample a normal observation dataset $\mathcal{D}_{normal}$ online. Then, we directly apply adversarial attacks to this dataset. Given an adversary $\mathcal{A}_{\mathcal{B}}$, we build an adversarial dataset $\mathcal{D}_{adv} = \{ \mathcal{A}_{\mathcal{B}}(o) \mid o \in \mathcal{D}_{normal} \}$. In Section~\ref{sec:attack}, we will demonstrate why two types of adversarial attacks can generate the $\mathcal{D}_{adv}$ without interacting with an environment under adversarial attacks.

\subsubsection{Robustness Regularizer}

A robustness regularizer~\cite{zhang2020robust} can also be integrated into our defense schema. The intuition behind the robustness regularizer is that if we can minimize the difference between the action distribution under normal observations and the action distribution under attacks, the robustness of our network can be improved. A robustness regularizer measures this difference.

Assuming a denoiser $\mathtt{den}$ and pretrained policy $\pi$, the action $a = \pi(\mathtt{den}(o))$. We treat $\pi$ and $\mathtt{den}$ as one network $\pi_{\mathtt{den}}$. Given an attack $\hat{o} = \mathcal{A}_{\mathcal{B}}(o)$ and the policy covariance matrix $\Sigma$, the robustness regularizer for stochastic PPO is
\begin{align*}
    R_{\text{ppo}} = \left(\pi_{\mathtt{den}}\left(\mathcal{A}_{\mathcal{B}}(o)\right)-\pi_{\mathtt{den}}\left(o\right)\right) \cdot \Sigma^{-1} \cdot \left(\pi_{\mathtt{den}}\left(\mathcal{A}_{\mathcal{B}}(o)\right)-\pi_{\mathtt{den}}\left(o\right)\right)
\end{align*}
and the robustness regularizer for deterministic TD3 is:
\begin{align*}
    R_{\text{td3}} = \left\| \pi_{\mathtt{den}}\left(\mathcal{A}_{\mathcal{B}}(o)\right)-\pi_{\mathtt{den}}\left(o\right) \right\|_{2}.
\end{align*}
Following~\cite{zhang2021robust}, the attack $\mathcal{A}_{\mathcal{B}}$ considered here is the opposite attack that will be introduced in Sec.~\ref{sec:attack}. The opposite attack depends on the policy network. When computing the robustness regularizer, we attack $\pi_{\mathtt{den}}$ instead of $\pi$. The theoretical foundation for minimizing the difference between action distributions is provided by Theorem 5 in~\cite{zhang2020robust}. It shows the total variance between the normal action distribution and the action distribution generated by observation $\hat{o}$ under attack can bound the value function (i.e., performance) difference. However, unlike \cite{zhang2021robust} that trains a \emph{policy} with a robustness regularizer, we achieve this by training the parameters of the \emph{denoiser} $\mathtt{den}$, and retain the parameters of the pretrained policy $\pi$.

The regularizers can be added with the denoiser's objective function directly. Then, according to the policy type, we optimize $L_{\mathtt{den}} + R_{\text{ppo}}$ or $L_{\mathtt{den}} + R_{\text{td3}}$ to update the denoiser's parameters. Optimizing the denoiser's objective function and robustness regularizer focus on different goals. 
A small value of $L_{\mathtt{den}}$ means the output of the denoiser is close to the groundtruth observations, while a small $R_{\text{ppo}}$ or $R_{\text{td3}}$ means the action distributions in adversarial and non-adversarial scenarios are similar.

\subsection{Observation Attacks}
\label{sec:attack} 

\subsubsection{Attacks}
We evaluate our defense on four well-studied categories of observation attacks. In this section, we briefly introduce these attacks and explain why the opposite attack and Q-function attack can be used to generate offline adversarial datasets without sampling under adversarial scenarios.

\paragraph{Opposite Attack}
The opposite attack appears in \cite{huang2017adversarial,pattanaik2017robust,zhang2020robust,lin2017tactics}. By perturbing observations, this attack either minimizes the likelihood of the action with the highest probability~\cite{huang2017adversarial,pattanaik2017robust,zhang2020robust} or maximizes the likelihood of the least-similar action~\cite{lin2017tactics} in discrete action domains. 
We choose to minimize the likelihood of the preferred action. The attacked observation $\hat{o}_t$ is computed as:
\begin{align}
    \hat{o}_t = \underset{\hat{o}_t \in \mathcal{B}(o_t)}{\mathtt{argmax}}\ l_{op}(o_t, \hat{o}_t),
    \label{eq:opposite-attack}
\end{align}
where $\mathcal{B}(o_t)$ signifies all the allowed perturbed observations around $o_t$. For stochastic policies, $ l_{op}(o_t, \hat{o}_t) = (\pi(o_t) - \pi(\hat{o}_t)) \Sigma^{-1} (\pi(o_t) - \pi(\hat{o}_t)), $ where $\pi(o_t)$ and $\pi(\hat{o}_t)$ are the mean of the predicted Gaussian distribution, and $\Sigma$ is the policy covariance matrix. For a deterministic TD3 algorithm, the difference is defined as the Euclidean distance between the predicted actions, $ l_{op}(o_t, \hat{o}_t) = ||\pi(o_t) - \pi(\hat{o}_t)||_2.  $ This attack only depends on the policy $\pi$. Given a normal dataset $\mathcal{D}_{normal}$, we can apply this attack on every observation in $\mathcal{D}_{normal}$ to generate the adversarial dataset $\mathcal{D}_{adv} = \{ \mathcal{A}_{\mathcal{B}}(o) \mid o \in \mathcal{D}_{normal} \}$ without any interaction with the environment. Since generating $\mathcal{D}_{normal}$ and applying the opposite attack does not sample under adversarial scenarios. Thus, generating $\mathcal{D}_{adv}$ does not require sampling under adversarial scenarios.

\paragraph{Q-function Attack}
\cite{kos2017delving,pattanaik2017robust,zhang2020robust} compute observation perturbations with the Q-function $Q(o_t, a_t)$. This attack only depends on the Q function $Q(o_t, a_t)$. The Q function sometimes comes with trained policies (e.g., TD3). When the Q function is not accompanied by trained policies (e.g, PPO), the Q-function can be learnt under non-adversarial scenarios~\cite{zhang2021robust}. We want to find a $\hat{o}_t$ such that it minimizes the $Q$ under budget $\mathcal{B}$. Thus, the attacked observation $\hat{o}_t$ is computed as
\begin{align}
    \hat{o}_t = \underset{\hat{o}_t \in \mathcal{B}(o_t)}{\mathtt{argmin}}\ Q(o_t, \pi(\hat{o}_t))
    \label{eq:q-function-attack}
\end{align}
We can generate the adversarial dataset $\mathcal{D}_{adv}$ with $\mathcal{D}_{normal}$ and $Q(o_t, a_t)$. Notice that getting $\mathcal{D}_{normal}$ and $Q(o_t, a_t)$ does not require interacting with environments under attacks. Therefore, the Q-function attack can also generate the $\mathcal{D}_{adv} = \{ \mathcal{A}_{\mathcal{B}}(o) \mid o \in \mathcal{D}_{normal} \}$ without sampling under adversarial scenarios.

\paragraph{Optimal Attack} 
The optimal attack learns an adversarial policy $\pi_{adv}$ adding perturbation $\Delta_{o_t}$ to the observation $o_t$. For example, \cite{zhang2021robust} demonstrated this strong attack over MuJoCo benchmarks. \cite{gleave2019adversarial} learns such an adversarial policy in two-player environments. The action outputted by the adversarial policy is $\Delta_{o_t}$, and the input of $\pi_{adv}$ is $o_t$. The perturbed observation is
\begin{align}
    \hat{o}_t = \mathtt{proj}_{\mathcal{B}}(o_t + \Delta_{o_t})
    \label{eq:optimal-attack}
\end{align}
where $\mathtt{proj}_{\mathcal{B}}$ is a projection function that constrains the perturbed observation $\hat{o}_t$ to satisfy the attack budget $\mathcal{B}$. The adversarial policy is trained to \emph{minimize} the cumulative discounted reward $\mathcal{R}$. Importantly, training this adversarial policy $\pi_{adv}$ requires adversarial sampling online. Thus, we did not adopt it to generate our adversarial dataset $\mathcal{D}_{adv}$.

\paragraph{Enchanting Attack}
This type of attack first appeared in \cite{lin2017tactics}. It integrates a planner into the attack loop. The planner generates a sequence of adversarial actions, and the adversary crafts perturbations to mislead neural network policies to output adversarial action sequences. At time step $t$, an adversarial motion planner generates a sequence of adversarial actions $[a_{t,0}, a_{t,1}, ..., a_{t,T-t}]$ guiding the agent to perform poorly. Since we attack the observation space and cannot change the action directly, we need to perturb observations to mislead the policy to predict the planner's adversarial actions. Given the policy network $\pi$, the perturbed observation is
\begin{align}
    \hat{o}_t = \underset{\hat{o}_t \in \mathcal{B}}{\mathtt{argmin}}\ ||\pi(\hat{o}_t) - a_{t,0}||_2
    \label{eq:enchanting-attack}
\end{align}
The $a_{t,0}$ is the target adversarial action. In our attack, we call the planner at every step and use the first action as an adversarial action, which avoids the errors caused by the deviation between the actual trajectory and planned trajectory, and thus strengthens the enchanting attack. For the continuous control problem, we use a Cross-Entropy Motion (CEM) planner~\cite{kobilarov2012cross} for adversarial planning. Generating or applying an adversarial planner typically requires online adversarial sampling. Therefore, we did not generate the $\mathcal{D}_{adv}$ with the enchanting attack.

To summarize, we evaluate our defense over four types of attacks. However, we only generate the adversarial dataset $\mathcal{D}_{adv}$ with the opposite attack and Q-function attack because they do not require risky online adversarial sampling. Sec.~\ref{sec:exp} shows that the denoiser trained with the adversarial dataset generated from these two attacks alone performs surprisingly well even when used in defense against all the four attacks we consider.

\subsubsection{When to Attack}

Since we want to minimize the reward with as few perturbations as possible, it is crucial to attack when the agent is vulnerable. We use the value function approximation as the indicator of vulnerability. When the value function predicts a certain observation has a small future value, such an observation is likely to cause a lower cumulative reward. A lower cumulative reward shows either the vulnerability of this observation itself (e.g., a running robot is about to fall) or the vulnerability of the corresponding policy (i.e., the policy would perform poorly given this observation). Thus, we can use the value function approximation to choose the time to trigger our attack. Given an observation $o_t$ and the value function $V$, by choosing a threshold $C_{vul}$, we only trigger the attack when $V(o_t) < C_{vul}$. The vulnerability indicator is $$\mathbbm{1}_{vul}(o_t) := V(o_t) < C_{vul}$$ We use the value function learned during training for the PPO policy. Because $V(o_t) = \int_{a_t \sim \pi(o_t)} Q(o_t, a_t)$ and $a_t = \pi(o_t)$ for a deterministic policy, $V(o_t) = Q(o_t, \pi(o_t))$. Hence, we can compute the value function of TD3 with the learned Q function and policy. We tune $C_{vul}$ to achieve the strongest attack while minimizing the number of perturbations triggered.

\section{Experiments}
\label{sec:exp}

We evaluate our approach on five continuous control tasks with respect to a stochastic PPO policy and a deterministic TD3 policy. The PPO policies were trained by ourselves, and the TD3 policies use pretrained models from \cite{rl-zoo3}. Our experiments answer the following questions.

\begin{itemize}
    \item[\textbf{Q1.}] Does our defense improve robustness against adversarial attacks?
    \item[\textbf{Q2.}] How does our defense impact performance in non-adversarial scenarios?
    \item[\textbf{Q3.}] How does our approach compare with state-of-the-art online adversarial training approach?
    \item[\textbf{Q4.}] How is the performance of our detectors and denoisers in terms of accuracy?
    \item[\textbf{Q5.}] How does our defense perform under adaptive attacks?
\end{itemize}

\subsection{Rewards under Attack w/wo Defense (Q1)}
\label{sec:attack-defense}

In this section, we show how our defense improves robustness. We report attack and defense results on the pretrained policies in Table~\ref{tab:attack-defense-rewards}. The ``Benchmark'' and ``Algo'' columns are the continuous control tasks and the reinforcement learning policies, resp. The ``Dimension'' column contains the dimensionality information of state and action space. The $\varepsilon$ column shows the $\varepsilon$ of attack budget $\mathcal{B}$. The $\varepsilon$ of ``Hopper'', ``HalfCheetah'', and ``Ant'' are the same as the attack budget provided in \cite{zhang2021robust}; we increased $\varepsilon$ in ``Walker2d'' to 0.1. The ``Humanoid'' with the highest observation and action dimension is not evaluated in \cite{zhang2021robust}. We choose $\varepsilon = 0.15$ for ``Humanoid''.

\begin{table}[!htp]\centering
    \vspace{-5px}
    \caption{Benchmark Information and Rewards under Attack w/wo Defense}\label{tab:attack-defense-rewards}
    \scriptsize
    \begin{tabular}{lr|c|cc|ccccc}\toprule
        \multirow{2}{*}{Benchmark}   & \multirow{2}{*}{Algo} & \multirow{2}{*}{$\varepsilon$} & \multicolumn{2}{c|}{Dimension} & \multicolumn{4}{c}{Attack/Defense}                                                   \\\cmidrule{4-9}
                                     &                       &                                & state                          & action                             & Opposite  & Q-function & Optimal   & Enchanting \\\midrule
        \multirow{2}{*}{Hopper}      & TD3                   & \multirow{2}{*}{0.075}         & \multirow{2}{*}{11}            & \multirow{2}{*}{3}                 & 390/2219  & 960/3328   & 267/2814  & 1629/3287  \\
                                     & PPO                   &                                &                                &                                    & 271/2615  & 700/3569   & 247/3068  & 217/2751   \\\midrule
        \multirow{2}{*}{Walker2d}    & TD3                   & \multirow{2}{*}{0.1}           & \multirow{2}{*}{17}            & \multirow{2}{*}{6}                 & 751/4005  & 478/4329   & 187/4772  & 762/4538   \\
                                     & PPO                   &                                &                                &                                    & 241/1785  & 3510/4737  & -38/1393  & 1582/1741  \\\midrule
        \multirow{2}{*}{HalfCheetah} & TD3                   & \multirow{2}{*}{0.15}          & \multirow{2}{*}{17}            & \multirow{2}{*}{6}                 & 1770/8946 & 1603/8471  & 1017/8174 & 1802/8838  \\
                                     & PPO                   &                                &                                &                                    & 1072/6115 & 1665/4218  & 833/3765  & 274/4477   \\\midrule
        \multirow{2}{*}{Ant}         & TD3                   & \multirow{2}{*}{0.15}          & \multirow{2}{*}{111}           & \multirow{2}{*}{8}                 & 603/3516  & -46/2137   & -893/2809 & 522/4729   \\
                                     & PPO                   &                                &                                &                                    & -351/5404 & -157/1042  & 558/4574  & 196/5497   \\\midrule
        \multirow{2}{*}{Humanoid}    & TD3                   & \multirow{2}{*}{0.15}          & \multirow{2}{*}{376}           & \multirow{2}{*}{17}                & 431/4849  & 454/4042   & 585/5130  & 420/5125   \\
                                     & PPO                   &                                &                                &                                    & 531/3161  & 406/3508   & 415/3630  & 396/1695   \\
        \bottomrule
    \end{tabular}
    \vspace{-5px}
\end{table}

We provide attack and defense results in the ``Attack/Defense'' column. The four sub-columns in this column are the attacks we described in Section~\ref{sec:attack}. The numbers before the slash are the cumulative rewards gained under attack. In this table, we assume the adversary is not aware of our defense's existence. The experiment results show that these strong attacks can significantly decrease the benchmarks' rewards, and our defense significantly improved rewards for all attacks.

\subsection{Non-adversarial Scenarios (Q2) and Comparison (Q3)}

We evaluate rewards in non-adversarial scenarios and compare them with ATLA \cite{zhang2021robust}, a state-of-the-art online adversarial training approach, in this section. Adversarial attacks do not always happen. Therefore, maintaining strong performance in normal cases is essential. The ``Non-adversarial'' column summarizes the reward gained by policies without any adversarial attack injected. Rewards are computed as the average reward over 100 rollouts. The ``Pre.'' column shows the cumulative reward of pretrained policies, while the ``ATLA'' column is the reward gained by the ATLA policy in \cite{zhang2021robust}. The ``Ours'' column is the reward gained by the policies under our defense. The numbers in parentheses are the percentages of rewards preserved when compared with the pretrained policies, which are computed with reward in ``Ours'' divided by reward in ``Pre.''. Observe that the introduction of the detector preserves the performance of pretrained policies. Because our defense only intervenes when it detects anomalies, it has a mild impact on the pretrained policies in non-adversarial cases. In contrast, ATLA policies do not perform as well as our defended policies when no adversary appears on all the benchmarks.

\begin{table}[ht]\centering
    \vspace{-5px}
    \caption{Rewards in Non-adversarial Scenarios and Comparison}\label{tab:attack-and-defense-of-pretrained}
    \scriptsize
    \begin{tabular}{ll|ccc|ccc}\toprule
        \multirow{2}{*}{Benchmark}   & \multirow{2}{*}{Algo} & \multicolumn{3}{c|}{Non-adversarial} & \multicolumn{2}{c}{Avg./Min (Best Attack)}                                                                 \\\cmidrule{3-7}
                                     &                       & Pre.                                 & ATLA                                       & Ours       & ATLA                            & Ours           \\\midrule
        \multirow{2}{*}{Hopper}      & TD3                   & 3607                                 & \multirow{2}{*}{3220}                      & 3506(0.97) & \multirow{2}{*}{2192/1761(opt)} & 2912/2219(ops) \\
                                     & PPO                   & 3206                                 &                                            & 3201(1.00) &                                 & 3001/2615(ops) \\\midrule
        \multirow{2}{*}{Walker2d}    & TD3                   & 4719                                 & \multirow{2}{*}{3819}                      & 4712(1.00) & \multirow{2}{*}{1988/1430(opt)} & 4411/4005(ops) \\
                                     & PPO                   & 4007                                 &                                            & 3980(0.99) &                                 & 2414/1393(opt) \\\midrule
        \multirow{2}{*}{HalfCheetah} & TD3                   & 9790                                 & \multirow{2}{*}{6294}                      & 8935(0.91) & \multirow{2}{*}{5104/4617(enc)} & 8607/8174(opt) \\
                                     & PPO                   & 8069                                 &                                            & 7634(0.95) &                                 & 4644/3765(opt) \\\midrule
        \multirow{2}{*}{Ant}         & TD3                   & 5805                                 & \multirow{2}{*}{5313}                      & 5804(1.00) & \multirow{2}{*}{4310/3765(q)}   & 3298/2137(q)   \\
                                     & PPO                   & 5698                                 &                                            & 5538(0.97) &                                 & 4129/1042(q)   \\\midrule
        \multirow{2}{*}{Humanoid}    & TD3                   & 5531                                 & \multirow{2}{*}{4108}                      & 5438(0.98) & \multirow{2}{*}{3311/2719(q)}   & 4786/4042(q)   \\
                                     & PPO                   & 4568                                 &                                            & 4429(0.97) &                                 & 2999/1695(enc) \\
        \bottomrule
    \end{tabular}
    \vspace{-5px}
\end{table}

The column ``Avg./Min.(Best Attack)'' show statistics comparing ATLA and our defense under the four attacks. The numbers before the slash are the average reward gained under attacks, and the numbers after the slash are the lowest rewards among all the attacks. The abbreviations in parentheses are the best attack that achieves the lowest reward, where ``ops'' means the opposite attack, ``q'' means the Q-function attack, ``opt'' means the optimal attack, and ``enc'' means the enchanting attack. The results show that our defense trained with data sampled under non-adversarial scenarios provides comparable results with the riskier online adversarial training approach. Observe that 6 out of 10 benchmarks have a higher reward than ATLA for the average rewards over attacks. For the worst rewards over attacks, our defense has a higher reward than the ATLA on 5 of 10 benchmarks. The result is surprising considering that we do not sample any adversarial observations online.

\subsection{Detector and Denoiser (Q4)}

The detector's performance is crucial for our defense since it prevents unnecessary interventions. We report the detectors' accuracy in non-adversarial scenarios and their F1 scores and false-negative rates under attack. The accuracy measures detectors' performance when no attack appears, and the F1 score measures how well the detectors perform when policies are under attack. Meanwhile, the false-negative rate tells us the percentage of adversarial attacks that are not detected. We present these results in the ``Detector'' column of Table~\ref{tab:detector-denoiser-scores}.

\begin{table}[!htp]\centering
    \vspace{-5px}
    \caption{Detector and Denoiser Performance}\label{tab:detector-denoiser-scores}
    \scriptsize
    \begin{tabular}{lr|c|cccc|cccc|ccccc}\toprule
        \multirow{3}{*}{Benchmark}   & \multirow{3}{*}{Algo} & \multicolumn{9}{c|}{Detector} & \multicolumn{4}{c}{Denoiser}                                                                                                                                                          \\\cmidrule{3-15}
                                     &                       & Acc.                          & \multicolumn{4}{c|}{F1 Score} & \multicolumn{4}{c|}{False Negative Rate} & \multicolumn{4}{c}{Mean Absolute Error}                                                                    \\\cmidrule{3-15}
                                     &                       & Normal                        & Ops                           & Q                                        & Opt                                     & Enc  & Ops  & Q    & Opt  & Enc  & Ops   & Q     & Opt   & Enc   \\\midrule
        \multirow{2}{*}{Hopper}      & TD3                   & 0.99                          & 0.82                          & 0.98                                     & 0.88                                    & 0.99 & 0.00 & 0.01 & 0.07 & 0.00 & 0.030 & 0.023 & 0.032 & 0.024 \\
                                     & PPO                   & 0.99                          & 0.97                          & 0.94                                     & 0.94                                    & 0.95 & 0.00 & 0.00 & 0.00 & 0.00 & 0.026 & 0.018 & 0.034 & 0.038 \\\midrule
        \multirow{2}{*}{Walker2d}    & TD3                   & 0.99                          & 0.99                          & 0.99                                     & 0.99                                    & 0.99 & 0.01 & 0.01 & 0.02 & 0.01 & 0.030 & 0.032 & 0.042 & 0.045 \\
                                     & PPO                   & 0.95                          & 0.97                          & 0.93                                     & 0.95                                    & 0.99 & 0.05 & 0.01 & 0.00 & 0.01 & 0.041 & 0.030 & 0.033 & 0.037 \\\midrule
        \multirow{2}{*}{HalfCheetah} & TD3                   & 0.95                          & 0.99                          & 0.98                                     & 0.98                                    & 0.96 & 0.00 & 0.00 & 0.00 & 0.00 & 0.049 & 0.048 & 0.050 & 0.043 \\
                                     & PPO                   & 0.99                          & 0.99                          & 0.98                                     & 0.96                                    & 0.96 & 0.00 & 0.01 & 0.02 & 0.01 & 0.057 & 0.041 & 0.046 & 0.048 \\\midrule
        \multirow{2}{*}{Ant}         & TD3                   & 0.99                          & 0.99                          & 0.99                                     & 0.99                                    & 0.99 & 0.00 & 0.00 & 0.00 & 0.00 & 0.022 & 0.022 & 0.022 & 0.023 \\
                                     & PPO                   & 0.99                          & 0.99                          & 0.99                                     & 0.99                                    & 0.99 & 0.00 & 0.00 & 0.00 & 0.00 & 0.023 & 0.024 & 0.027 & 0.026 \\\midrule
        \multirow{2}{*}{Humanoid}    & TD3                   & 0.99                          & 0.96                          & 0.97                                     & 1.00                                    & 0.96 & 0.08 & 0.04 & 0.00 & 0.07 & 0.048 & 0.047 & 0.043 & 0.046 \\
                                     & PPO                   & 0.99                          & 0.99                          & 0.99                                     & 0.99                                    & 0.99 & 0.00 & 0.00 & 0.00 & 0.00 & 0.055 & 0.045 & 0.048 & 0.050 \\
        \bottomrule
    \end{tabular}
    \vspace{-5px}
\end{table}

The detector is expected only to report negative in non-adversarial scenarios. Since there is no adversarial observation (i.e., positive sample) in non-adversarial scenarios, we measure the detector's quality with accuracy instead of the F1 score. The 3rd column in Table~\ref{tab:detector-denoiser-scores} reports the accuracy of all the detectors in non-adversarial scenarios. The worst accuracy is $0.95$. The high accuracy explains why our defense retains the performance of pretrained policies in non-adversarial cases. We measure the quality of detectors under attack with the F1 scores and false-negative rate. When attacking Hopper's PPO policy with the opposite attack and optimal attack, the F1 scores are $0.82$ and $0.88$, respectively. However, their false negative rates are $0.00$ and $0.01$, respectively. The low false-negative rates show that our detectors ensure the denoiser would be triggered under attack. Moreover, the data shows that the relatively low F1 score was caused by false positives, which means the defense will be cautious and use denoised observations more often. The left data has an F1 score higher than $0.94$ and a false negative rate lower than $0.04$, which supports our claim that the detector works well when the policies are under attack.

The Mean Absolute Errors (MAEs) between the outputs of the denoiser and groundtruth observations are reported in the ``Denoiser'' column in Table~\ref{tab:detector-denoiser-scores}. Although we only train the denoiser with the augmented data generated with the opposite and Q-function attack, the MAE of the optimal attack and enchanting attack is close to the MAE of the opposite attack and Q-function attack. This explains why our defense also works well on the opposite and enchanting attacks, as shown in Table~\ref{tab:attack-defense-rewards}.

\subsection{Adaptive Attack (Q5)}

We further evaluated the robustness of our defense under adaptive attacks. The defense in Section~\ref{sec:attack-defense} is evaluated when the attacks are not aware of the existence of our defense. However, once the adversaries realize that we have upgraded our defense, they can jointly attack our defense and pretrained policies. When the adversary can access both the detector and denoiser, it can mislead the detector to ignore anomalies with adversarial observations. We briefly introduce the key idea of adaptive attacks here. A more formal description of our adaptive attack design is provided in Appendix B.

The adversary needs to attack our defense and the pretrained policy jointly. Firstly, we consider how to attack the denoiser. Under our defense, the action $a_t$ is computed with a sequential model $a_t = \pi(\mathtt{den}(o_t))$;  we thus replace the pretrained policy $\pi(o_t)$ with $\pi(\mathtt{den}(o_t))$ and attack this sequential model. Secondly, adaptive attacks also need to fool the detector. Because the anomaly is defined with respect to being greater than a threshold, a malicious observation should decrease the $\ell_\infty$-norm in $\mathbbm{1}_{anomaly}$. This objective can be defined with a loss term $l_{det}(o_t) = || \mathtt{det}(o_t) - o_t ||_\infty$.  For the opposite attack, q-function attack, and enchanting attack, in addition to using $\pi(\mathtt{den}(o_t))$ to replace $\pi(o_t)$, we optimize $l_{det}(o_t)$ jointly with Eq.~\eqref{eq:opposite-attack}, Eq.~\eqref{eq:q-function-attack}, and Eq.~\eqref{eq:enchanting-attack} respectively. For the optimal attack, we train the adversarial policy with the involvement of our defense.

\begin{table}[!htp]\centering
    \vspace{-5px}
    \caption{Adaptive Attack (\% change in reward)}\label{tab:adaptive-attack}
    \scriptsize
    \begin{tabular}{lr|rrrr|rrr}\toprule
        Benchmark                  & Algo & Ops   & Q     & Opt   & Enc   & Min   & Max   \\\midrule
        \multirow{2}{*}{Hopper}      & TD3    & 0.63  & 0.06  & 0.17  & 0.04  & 0.04  & 0.63  \\
                                     & PPO    & -0.16 & -0.18 & -0.12 & -0.11 & -0.18 & -0.11 \\\midrule
        \multirow{2}{*}{Walker2d}    & TD3    & 0.19  & 0.10  & 0.00  & 0.05  & 0.00  & 0.19  \\
                                     & PPO    & 0.12  & -0.18 & -0.16 & 0.17  & -0.18 & 0.17  \\\midrule
        \multirow{2}{*}{HalfCheetah} & TD3    & -0.14 & 0.06  & 0.08  & 0.03  & -0.14 & 0.08  \\
                                     & PPO    & -0.23 & 0.10  & 0.17  & 0.20  & -0.23 & 0.20  \\\midrule
        \multirow{2}{*}{Ant}         & TD3    & -0.28 & -0.17 & -0.14 & -0.18 & -0.28 & -0.14 \\
                                     & PPO    & -0.24 & 0.40  & 0.11  & -0.22 & -0.24 & 0.40  \\\midrule
        \multirow{2}{*}{Humanoid}    & TD3    & 0.09  & 0.28  & -0.03 & -0.06 & -0.06 & 0.28  \\
                                     & PPO    & 0.28  & 0.23  & 0.01  & 0.06  & 0.01  & 0.28  \\
        \bottomrule
    \end{tabular}
    \vspace{-5px}
\end{table}

We use the defense rewards in Table \ref{tab:attack-defense-rewards} (numbers after the slash) as baselines and report the percentages by which reward changes under adaptive attacks in Table~\ref{tab:adaptive-attack}. The benchmark column contains the task names and the policy types. We have introduced the attack name abbreviations in Section~\ref{sec:attack-defense}, and the rewards changes under these attacks are reported from column 2 to column 5. The ``Min'' and ``Max'' columns are the minimal and maximal changes comparable with the baseline rewards. In the worst case, the adaptive attack causes the performance on Ant-TD3 to decrease $28\%$ under the opposite attack. We can observe that some rewards increase under the adaptive attack. This is because jointly attacking the detector can be challenging for the adversary. Since the detector is also a GRU-VAE, the first problem the adversary needs to address is the stochasticity introduced by the detector and denoiser themselves. Moreover, the adversary needs to fool the policy and detector simultaneously, which increases the difficulty of attacking our defense.

\section{Conclusion}

This paper proposes a detect-and-denoise defense against the observation attacks on deep reinforcement learning. Our defense samples the adversarial observations offline and thus avoids the risky online sampling under adversarial attacks. In the absence of an adversary, our defense does not compromise performance. We evaluated our approach over four strong attacks with five continuous control tasks under both stochastic and deterministic policies. Experiment results show that our approach is comparable to previous online adversarial training approaches, provides reasonable performance under adaptive attacks, and does not sacrifice performance in normal (non-adversarial) settings.

\bibliographystyle{splncs04}
\bibliography{ecml22}

\newpage
\appendix
\newpage

\section{Performance Bound}
We prove the performance bound for Gaussian stochastic policy with constant independent variance and deterministic policy on fully-observable MDP. Given a policy $\pi$, value functions $V_{\pi}(o)$ is the cumulative discounted future reward of the observation $o$. We care about the performance changes before and after attacks. The performance changes can be measured by the value function difference between the pretrained policy and the policy under attack.

Theorem 5 in \cite{zhang2020robust} provides an upper bound on the max difference between value functions with different action distributions, formally,
\begin{align}
    \label{eq:zhang-theorem-5}
    \max_{o\in\mathcal{S}}\big\{V_\pi(o) - V_{\pi}(o') \big\}\leq \alpha \max _{o \in S} \max _{o' \in
        \mathcal{B}(o)} \mathrm{D}_{\mathrm{TV}}(\pi(\cdot \mid o), \pi(\cdot \mid o'))
\end{align}
where $\mathrm{D}_{\mathrm{TV}}(\pi(\cdot \mid o), \pi(\cdot \mid o'))$ is the total variation distance between $\pi(\cdot \mid o)$ and $\pi(\cdot \mid o')$, and $\alpha$ is a constant that does not depend on $\pi$. Assuming that the policy network is Lipschitz continuous, we show that as the denoiser accuracy improves, the difference between value functions reduces.

\begin{theorem}
    \label{thm:optimal_distance}
    Given a Gaussian stochastic policy with constant independent variance $\pi$ and its value function $V_\pi(o)$, assuming that the policy is Lipschitz continuous, for all $o \in O$ we have
    \begin{equation}
        \label{eq:difference_tv}
        \begin{aligned}
            \max_{o\in\mathcal{S}}\big\{{V}_\pi(o) - {V_{\pi}}(\hat{o}) \big\}\leq \beta \max_{o\in O}\max_{o' \in \mathtt{den}(\mathcal{B}(o))} || \mathtt{den}(\hat{o}) - o||_2
        \end{aligned}
    \end{equation}
    where $ \mathtt{den} :O \rightarrow O$ is the denoiser, $\hat{o} = \mathcal{A}_\mathcal{B}(o)$ and
    $\beta$ is a constant that does not depend on $\pi$.
\end{theorem}

Theorem~\ref{thm:optimal_distance} tells us that the performance difference before and under attack is bounded by the max Euclidean difference between the normal observation and the output of the denoiser. A more accurate denoiser gives a tighter upper bound, and thus better preserves pretrained policy's performance.

\begin{proof}
    From Pinsker's inequality,
    \begin{align}
        \label{eq:var-kl}
        \mathrm{D}_{TV}(\pi(\cdot|o),\pi(\cdot|o^\prime)) \leq \sqrt{\frac{1}{2}\mathrm{D}_{KL}(\pi(\cdot|o)||\pi(\cdot|o^\prime))}
    \end{align}

    We assume that our stochastic policies follow a Gaussian distribution with constant diagonal covariance matrix. Supposing that $\pi(\cdot|o)$'s mean is $\mu_1$ and covariance matrix is $\Sigma_1$, and $\pi(\cdot|o^\prime)$'s mean is $\mu_2$ and covariance matrix is $\Sigma_2$; $o, o' \in \mathbb{R}^d$,
    \begin{align*}
        \mathrm{D}_{KL} & (\pi(\cdot|o)||\pi(\cdot|o^\prime))                                                                                                                                                                                                            \\
                        & = \frac{1}{2}\left(\log \frac{\left|\Sigma_{2}\right|}{\left|\Sigma_{1}\right|}-d+\operatorname{tr}\left\{\Sigma_{2}^{-1} \Sigma_{1}\right\}\right) + \frac{1}{2} \left(\mu_{2}-\mu_{1}\right)^{T} \Sigma_{2}^{-1}\left(\mu_{2}-\mu_{1}\right)
    \end{align*}

    Since $\Sigma_1$ and $\Sigma_2$ are constant matrices, $\frac{1}{2}\left(\log \frac{\left|\Sigma_{2}\right|}{\left|\Sigma_{1}\right|}-d+\operatorname{tr}\left\{\Sigma_{2}^{-1} \Sigma_{1}\right\}\right)$ is also a constant. Thus, there must exist a constant $C_1 \in \mathbb{R}^+$ such that $\forall \mu_1, \mu_2$,
    \begin{align*}
        \frac{1}{2} & \left(\log \frac{\left|\Sigma_{2}\right|}{\left|\Sigma_{1}\right|}-d+\operatorname{tr}\left\{\Sigma_{2}^{-1} \Sigma_{1}\right\}\right) + \frac{1}{2} \left(\mu_{2}-\mu_{1}\right)^{T} \Sigma_{2}^{-1}\left(\mu_{2}-\mu_{1}\right) \\
                    & \leq C_1 \left(\mu_{2}-\mu_{1}\right)^{T} \Sigma_{2}^{-1}\left(\mu_{2}-\mu_{1}\right)
    \end{align*}
    All the elements in $\Sigma_{2}^{-1}$ is positive, so there must exist $C_2 \in \mathbb{R}^+$ such that $\forall \mu_1, \mu_2$,
    \begin{align*}
        \left(\mu_{2}-\mu_{1}\right)^{T} \Sigma_{2}^{-1}\left(\mu_{2}-\mu_{1}\right) \leq C_2 \left\|\mu_{2}-\mu_{1}\right\|_2^2
    \end{align*}
    Thus,
    \begin{equation}
        \begin{aligned}
            \label{eq:kl-gaussian}
            \mathrm{D}_{KL} (\pi(\cdot|o)||\pi(\cdot|o^\prime)) & \leq C_1 \left(\mu_{2}-\mu_{1}\right)^{T} \Sigma_{2}^{-1}\left(\mu_{2}-\mu_{1}\right) \\
                                                                & \leq C_1 C_2 \left\|\mu_{2}-\mu_{1}\right\|_2^2
        \end{aligned}
    \end{equation}
    Now assuming our policy network is Lipschitz bounded, we have a constant $L$ such that
    \begin{equation}
        \begin{aligned}
            \label{eq:lipschitz}
            ||\mu_2 - \mu_1||_2 \leq L ||o' - o||_2
        \end{aligned}
    \end{equation}
    When integrating our denoiser, $o' = \mathtt{den}(\hat{o})$ and $o' \in \mathtt{den}(\mathcal{B}(o))$; combining \eqref{eq:zhang-theorem-5}, \eqref{eq:var-kl}, \eqref{eq:kl-gaussian}, and \eqref{eq:lipschitz}, we get
    \begin{equation}
        \begin{aligned}
            \max_{o\in\mathcal{S}}\big\{V_\pi(o) - V_{\pi}(\mathcal{A}_{\mathcal{B}}(o)) \big\} & \leq \alpha \max_{o\in O}\max_{o' \in \mathtt{den}(\mathcal{B}(o))} \sqrt{\frac{1}{2}\mathrm{D}_{KL}(\pi(\cdot|o)||\pi(\cdot|o^\prime))} \\
                                                                                                & \leq \alpha \max_{o\in O}\max_{o' \in \mathtt{den}(\mathcal{B}(o))} \sqrt{\frac{1}{2} C_1 C_2 \left\|\mu_{2}-\mu_{1}\right\|_2^2}        \\
                                                                                                & = \alpha \max_{o\in O}\max_{o' \in \mathtt{den}(\mathcal{B}(o))} \sqrt{\frac{1}{2} C_1 C_2 } \left\|\mu_{2}-\mu_{1}\right\|_2            \\
                                                                                                & \leq \alpha \max_{o\in O}\max_{o' \in \mathtt{den}(\mathcal{B}(o))} \sqrt{\frac{1}{2} C_1 C_2 } L \left\|o'-o\right\|_2                  \\
                                                                                                & = \alpha L \sqrt{\frac{1}{2} C_1 C_2 } \max_{o\in O}\max_{\hat{o} \in \mathcal{B}(o)} \|\mathtt{den}(\hat{o}) - o\|_2                    \\
                                                                                                & = \beta \max_{o\in O}\max_{o' \in \mathcal{B}(o)} \| \mathtt{den}(\hat{o}) - o\|_2
        \end{aligned}
    \end{equation}
    where $\beta = \alpha L \sqrt{\frac{1}{2} C_1 C_2 }.$

    For deterministic policy, we can add an independent Gaussian noise around its action (i.e., using the predicted action as the mean of a Gaussian distribution) and gain the same results.
    \qed
\end{proof}

This proof tells us that the accuracy of denoiser bounds the performance difference between adversarial and non-adversarial scenarios. Additionally, we want to point out that the \eqref{eq:zhang-theorem-5} supports why the robustness regularizer works. Optimizing the robustness regularizer reduces the distance between the action distributions of pretrained policy and when it is under attack. However, this is achieved by training the denoiser, which preserved the parameters of pretrained policy.

\section{Adaptive Attack Details}
\label{sec:adaptive-attack-details}
Our defense mechanism can also be the victim of all the four types of attacks. The adaptive attacks focus on two folders. First, they need to fool the detector so that the detector fails to alter attacked observations. This can be achieved by maximizing $l_{det}(o_t, \hat{o}_t)$,
\begin{align*}
    l_{det}(o_t, \hat{o}_t) = || \mathtt{det}(\hat{o}_t) - o_t ||_\infty.
\end{align*}
Second, the adaptive attacks should consider both the denoiser and the pretrained policy jointly to bypass the effects introduced by our denoiser. In other words, the adaptive attack needs to attack the sequential model $\pi_\mathtt{den}$ instead of the pretrained policy $\pi$.

\subsection{Opposite Adaptive Attack}
The opposite attack maximizes the distance between action distribution in non-adversarial and adversarial scenarios. When considering the adaptive attacks, we compute the attacked observation $\hat{o}$ with \eqref{eq:adaptive-ops-attack}.
\begin{align}
    \label{eq:adaptive-ops-attack}
    \hat{o}_t = \underset{\hat{o}_t \in \mathcal{B}(o_t)}{\mathtt{argmax}}\ \left( l'_{op}(o_t, \hat{o}_t) + l_{det}(o_t, \hat{o}_t) \right),
\end{align}
For Gaussian stochastic policy,
\begin{align*}
    l'_{op}(o_t, \hat{o}_t) = (\pi_\mathtt{den}(o_t) - \pi_\mathtt{den}(\hat{o}_t)) \Sigma^{-1} (\pi_\mathtt{den}(o_t) - \pi_\mathtt{den}(\hat{o}_t)),
\end{align*}
and for deterministic policy,
\begin{align*}
    l_{op}(o_t, \hat{o}_t) =  \|\pi_\mathtt{den}(o_t) - \pi_\mathtt{den}(\hat{o}_t)\|_2.
\end{align*}

\subsection{Q-function Adaptive Attack}
The Q-function attack minimizes the Q-function prediction in non-adversarial and adversarial scenarios. We compute the attacked observation $\hat{o}$ with \eqref{eq:adaptive-q-attack} for adaptive attacks.
\begin{align}
    \label{eq:adaptive-q-attack}
    \hat{o}_t = \underset{\hat{o}_t \in \mathcal{B}(o_t)}{\mathtt{argmin}}\ \left(Q(o_t, \pi_{\mathtt{den}}(\hat{o}_t)) - l_{det}(o_t, \hat{o}_t) \right)
\end{align}

\subsection{Optimal Adaptive Attack}
The optimal attack requires an adversarial policy to compute the perturbations added to normal observations $\Delta_{o_t} = \pi_{adv}(o_t)$.
\begin{align}
    \label{eq:adaptive-opt-attack}
    \hat{o}_t = \mathtt{proj}_{\mathcal{B}}(o_t + \Delta_{o_t})
\end{align}
For adaptive attack, the adversarial policy $\pi_{adv}$ is trained when the victim policies are augmented with our defense. The detailed training pipeline for an adversarial policy can be found in \cite{zhang2021robust}.

\subsection{Enchanting Adaptive Attack}
In step $t$, we generate the $a_{t,0}$ with an adversary CEM planner, and compute the adversarial observation with \eqref{eq:adaptive-enc-attack}.
\begin{align}
    \label{eq:adaptive-enc-attack}
    \hat{o}_t = \underset{\hat{o}_t \in \mathcal{B}}{\mathtt{argmin}}\ \left(||\pi_{\mathtt{den}}(\hat{o}_t) - a_{t,0}||_2 - l_{det}(o_t, \hat{o}_t) \right)
\end{align}

\section{Experiment Settings and Hyperparameters}
\label{sec:experiment-details}

\subsection{Dataset Size}
The normal trajectory dataset of Hopper, Walker2d, and HalfCheetah has $10,000$ trajectories, and each trajectory has $1,000$ observations. Ant, Humanoid's datasets have $20,000$ trajectories, and each trajectory has $1,000$ observations. When training the denoiser, we generate $\mathcal{D}_{adv}$ with the same number of trajectories via the opposite attack and Q-function attack on each benchmark.

\subsection{$C_{anomaly}$ and $C_{vul}$}
$C_{anomaly}$ decides how sensitive the detectors are. We tuned $C_{anomaly}$ to ensure that the false-negative rate is close to 0 and then make the F1 score as high as possible. The $C_{anomaly}$ can vary for each training on a detector, but it is easy to tune $C_{anomaly}$ with a simple linear search or Bayesian optimization in a short time. 

$C_{vul}$ controls the frequency of attack. We hope to downgrade the performance with the low attack frequency. Thus, we tuned $C_{vul}$ to ensure the performance decreased as significantly as \cite{zhang2021robust} while the attack frequency was as low as possible. The tuning process can also be done with a simple linear search or Bayesian optimization in a short time.  

\subsection{Detector and Denoiser Hyperparameters}

\begin{table}[!htp]\centering
    \caption{Detector and Denoiser Hyperparameters}\label{tab:det-den-hyper}
    \scriptsize
    \begin{tabular}{l|cccccc}\toprule
                            & Hopper                                 & Walker2d & HalfCheetah & Ant & Humanoid \\\midrule
        Encoder Hidden Size & 64                                     & 64       & 64          & 256 & 256      \\
        Encoder \#. Layer   & 1                                      & 1        & 1           & 1   & 2        \\\midrule
        Decoder Hidden Size & 64                                     & 64       & 64          & 256 & 256      \\
        Decoder \#. Layer   & 1                                      & 1        & 1           & 1   & 2        \\\midrule
        Embedding Size      & 64                                     & 64       & 64          & 64  & 128      \\\midrule
        Batch Size          & 128                                    & 128      & 128         & 256 & 1024     \\
        Epoch               & 50                                     & 50       & 50          & 100 & 200      \\
        Optimizer           & \multicolumn{5}{c}{Adam}                                                         \\
        Learning Rate       & \multicolumn{5}{c}{$1 \times 10^{-3}$}                                           \\
        \bottomrule
    \end{tabular}
\end{table}

We report the hyperparameters of our detector and denoiser in Table~\ref{tab:det-den-hyper}. The detector and denoiser on the same benchmark share the same hyperparameters.

Both the encoder and decoder are GRU. Thus, we report their ``Hidden Size'' and ``Encoder/Decoder \#. Layers'' (Number of Layers) in Table~\ref{tab:det-den-hyper}. The ``Embedding Size'' is the size of the latent space of a GRU-VAE. ``Batch Size'' and ``Epoch'' are the batch size and epoch when training detector and denoiser. For all detector and denoiser, the optimizer training the denoiser and detector is Adam, and the learning rate is $1 \times 10^{-3}$.

\end{document}